# Video Intelligence as a component of a Global Security system

Dominique Patrick VERDEJO[1], Eunika MERCIER-LAURENT[2]

[1] Personal Interactor, Montpellier and INHESJ, Paris, France,
,
[2] University Reims Champagne Ardenne, France

```
dominique.verdejo@personalinteractor.eu
eunika.mercier-laurent@univ-reims.fr
```

**Abstract.** This paper describes the evolution of our research from video analytics to a global security system with focus on the video surveillance component. Indeed video surveillance has evolved from a commodity security tool up to the most efficient way of tracking perpetrators when terrorism hits our modern urban centers. As number of cameras soars, one could expect the system to leverage the huge amount of data carried through the video streams to provide fast access to video evidences, actionable intelligence for monitoring real-time events and enabling predictive capacities to assist operators in their surveillance tasks. This research explores a hybrid platform for video intelligence capture, automated data extraction, supervised Machine Learning for intelligently assisted urban video surveillance; Extension to other components of a global security system are discussed.

Applying Knowledge Management principles in this research helps with deep problem understanding and facilitates the implementation of efficient information and experience sharing decision support systems providing assistance to people on the field as well as in operations centers. The originality of this work is also the creation of "common" human-machine and machine to machine language and a security ontology.

## 1 Introduction

Global digitalization has provided an extraordinary progress in communication and services with their downside, the rise of threats in the ever growing cyber domain. Increased terrorists' attacks and the rise of extremism require better management of all available resources in aim to detect planned actions and prevent disasters [30].

The complexity of security imposes considering all components and their interactions for better understanding of causes. The third hype of Artificial Intelligence has introduced powerful techniques for processing big data, used mainly in marketing at the beginning. Deep learning demonstrated great capacity in data processing; however the complex challenges at stake require combining both data and knowledge processing in hybrid systems. Knowledge Management approach helps deep understanding of security problem and impacts, organizing and optimizing all sources, such as video records, exchanges in social networks, data collected from IoT, sensors and others in aim to set an adequate architecture of decision support system for all participants. Various AI techniques allows, among others, verifying consistency of gathered information and detect fake news for instance. AI also support the whole knowledge flow generated by all involved actors.

Global security deals with security of persons, of buildings, but also cybersecurity of IT systems by early detection of threats and weak signals in the cyberspace. This paper gives an overview of all components and focus on enhancing the human-machine effectiveness of surveillance activities, taking video surveillance as primary application field.

Besides traffic and cleanliness control, the role of cameras in urban areas and public transportation systems is mostly to deter crime, theft and vandalism. This is achieved through two distinct activities, one being real-time and the second, post-event. In real-time, the operators in urban control centers apply techniques to follow individuals of interest or to monitor specific areas or persons to protect from attacks. Those activities are often carried out in close collaboration with police staff.

Investigation of video records a posteriori consists in analyzing them to locate the meaningful footages that can be used as evidences in a court or intelligence to track perpetrators. These post-event video investigations are often long and fastidious, but prove more and more efficient to identify the perpetrators and lead to their arrest as image quality and resolution steadily improve. In both contexts, the tremendous increase of the number of cameras represents a major challenge for the overall efficiency of the system. It has been demonstrated that a human operator can monitor 16 cameras over a period of 20 minutes [3]. Whichever activity, real-time or post-event, requires attention of the operator on a number of video feeds. This highlights the need for computer based operator assistance.

This problem was already studied in our previous research - Vortex decision support system for video analysis [34]. The current research aims at improving the previous system by applying Knowledge Management principles and integrating a feedback from the previous work.

To predict possible threat scenarios, this research focuses, among others, on Machine Learning models to process specific pattern matching in spatiotemporal event sequences. However the prediction based on past data is insufficient to predict the future disasters. The permanent metamorphosis of threats prevents from constructing a sufficiently representative and properly documented data set for system training.

Surveillance activities using video, social media and cyber defense data have one common point: they all capture and manipulate highly varied, complex and unpredictable events and behaviors, for which risk assessment requires mostly human common sense and contextual knowledge. Connection with the other sources of information and knowledge, such as the law enforcement files in the case of terrorism is necessary for overall understanding.

While related research focuses on media analysis, like video analysis, providing added forensic evidence research capabilities, we have chosen to emphasize on the analysis of metadata generated by such algorithms. These metadata can be efficiently used to establish correlations with other media analysis results like social networks or events generated by cyber security monitoring systems.

The applied bottom-up Knowledge Management approach combined with knowledge modelling allows easy extension to the others system components of global security, i.e. social networks analysis and cybersecurity, especially in the era of ubiquitous IoT and drones equipped with facial recognition capabilities.

After the general introduction and pointing out the challenge, this paper describes our research method, followed by the state of the art in the fields covered. The explanation on the role of Knowledge Management in this research is provided before we focus on the first system component of this research -hybrid decision support system for real-time video analysis. Conclusions and some perspectives of future work and applications are given at the end.

## 2      Research method

Considering our experience acquired during the initial work on Vortex project we decided to consider the video analysis as a part of a Global Security system. Currently, the proposed research method applies Knowledge Management and complex problem solving [35] to global security.  It consists in understanding the challenge of a global security system and aims at i) proposing a modular architecture for a hybrid system, ii) prototype and test separate modules conceived by combining holistic and system approaches. Each component plays its own role and interacts with the others for more accurate result. Incremental approach allows building the whole system step by step. At each stage, the adequate AI techniques are selected and tested and results discussed.
The role of Knowledge Management consists in

- Problem analysis with a given context. What data/information/knowledge do we need to successfully manage security and prevent disasters provoked by ill-intentioned people? What sources of data/information/knowledge are available?
- Initiating and managing a human-machine self-reinforcing synergy
- Defining the overall architecture of the decision support (not making) system, considering all involved actors' contributions and needs.

- Building generic and reusable knowledge models (static, dynamic or hybrid, ex. multi-agents systems).
- Defining a strategy for future extensions (Building blocs)
- Integrating a feedback on all stages.

Related to video analysis, we propose to transpose the general scene analysis problem into a natural language description exercise. This approach allows human operators to share understanding of the scene/situation with the machine. This common language, scene description capability, creates a gate between human and machine providing two major benefits: i) situational awareness knowledge can be immediately shared between operators without visual support and ii) the human operators can enrich the description provided by the machine with additional common sense or contextual knowledge.

This natural language processing approach will use ontologies as a model for text generation and a semantic navigation tool. This provides guidelines to keep consistency, manage ambiguities and act as a communication protocol dedicated to a specific media analysis.

The machine learns text generation through supervised machine learning (labeling) approach and builds risk scenarios using reinforced learning. The scenarios of maximum risk likelihood will be proposed to the operators for facing the most urgent challenge in modern control centers: monitoring exponentially growing numbers of sensors inputs.

This module will be considered using modularity, genericity and reusability approach for future extensions.

## 3 State of the art

Research in video analysis is not new. French programs PRIAM and RIAM initiated by the French Ministry of Industry gave some promising results in applying AI techniques for video analysis on the flow. For example, the architecture of MediaWorks combined neural networks, multi-agent systems and natural language processing for relevant finding on the flow of person or object to illustrate TV news [4]. These projects did not address security issues as the subject was not as critical at this time.

Since 2000, many researches have been made in semantic video indexing [4] and European Research has been funded to create tools to annotate and retrieve video [5]. It is commonly agreed today that we need an abstract layer of representation, a language, to describe and retrieve video. Ontologies have been proposed as an adapted tool to capture observations but also to shift domains as surveillance can be operated on media from many different natures depending on the activity [6][7][8] (satellite, urban, cyber, social networks, etc.)

The "Smart City" trend combining technological progress, cameras price drop down and raise of delinquency, has prompted cities to install surveillance systems. However they did not have a resource for systematic analysis of the recorded content. The raise of terrorist's attacks amplified a demand for efficient video surveillance systems.

The effectiveness of such systems depends on intelligence in exploring the available data sources of information and knowledge.
The analysis after the disaster pointed out the need for preventive actions.
Opening of public data and Big Data from multiple sources are available but using only analytics on the past data is insufficient for prevention of future disasters. We did not find in available literature the efficient decision support system to face this challenge. Some separate modules, such as video analytics, social networks mining, OSINT (Open Source Intelligence) and IoT data exploring are only described.

### 3.1 evolution of video surveillance

Learning from human generated annotations, the machine has shown a capacity to generalize, identifying objects and generating text sentences describing image and scenes. In 2017, this effort is extended to video with the DAVIS [14] challenge on video object segmentation. The downside of this approach, is that human contribution is highly necessary to generate meaningful datasets. Initiatives to deploy crowd-annotations platform have been recently undertaken to improve and speed-up the ground truth collection from users on the Web [15], fostering the need for creation of machine learning datasets.
It is notable that image segmentation, object recognition, video labeling creates a leap frog in surveillance applications and human interface. But, as every camera turns into a talkative witness, we lack the ability to correlate the different sources and enable global risk analysis. This is particularly important for behaviour analysis and anomaly detection in scene understanding, including gesture and intentions analysis.

### 3.2 Evolution and revolution of video analytics

Over the past 15 years, numerous tests and benchmarks were undertaken to assess feasibility of using algorithms to perform recognition of specific situations to ease the task of video operators. It is expected that automating video monitoring can lead to a less heavy mental workload for operators as their attention can be focused only on identified problems. In fact, false alarms tend to overcrowd the video monitoring environment and have rendered those technologies quite useless in most operational cases, specifically in large urban surveillance contexts.

Traditional video analytics, based on bitmap analysis can be useful to identify line crossing or counter-flow. They can count individuals and detect crowds and abandoned luggage. But they fail providing insights on more complex situations like fights, tagging, thefts and carjacking where more context and common sense is required [10]. Highly focused European FP7 2010-2013 research project, VANAHEIM ][9] has been developed in the context of two metro stations and revealed the difficulty to use inputs from video analytics modules to automate the displays on video walls. Nevertheless this project has been pivotal in demonstrating the huge potential of unsupervised video analysis and the detection of abnormality into long recordings.

Since 2010, a revolution has begun in video analytics. Thanks to Convolutional Neural Networks, Deep Learning techniques, object recognition, image segmentation and labeling has shown impressively efficient, up to the point where the machine, using a software built on top of GoogLeNet has demonstrated in 2015 an ability to identify objects in still images that is almost identical to humans[12]. This was made possible thanks to the availability of a very large image dataset called Imagenet [10](over 1,4 million images from over 1000 classes) manually annotated and a challenge that took place annually between 2010 and 2015, the ImageNet Large Scale Visual Recognition Challenge (ILSVRC). In 2015, the Chinese company BAIDU also claimed actual superiority of machine image recognition compared to human on the same image dataset [11].

### 3.3 Evolution of Cyber Security and Cyber Defense

Information systems security has evolved since 2000 taking into account the global interconnection of machines. System protection has evolved from basic anti-virus and firewalls up to a range of monitoring activities dedicated to detect and identify intrusions and potentially harmful activities on the networks and servers. Recently, companies have marketed AI based systems to achieve specific surveillance analysis, based on the logs produced by applications and systems. Interestingly, those new approaches, gathered around the SIEM components (Security Incident Event Management) tend to procure a behavioral analysis rather than a simple event categorization.

It is worth mentioning the similarities between video and cyber analytics which both converge toward behavioral analysis for detecting the weak signals in the massive data sets.

Cyber Security for instance, uses dedicated event description and exchange named STIX [32] to communicate cyber threat intelligence.

### 3.4 Evolution of Social Network Analysis

In complex situations such as terrorism and cybercrime analysis of video is not sufficient. Social networks and especially ephemeral ones are used to spread information and to recruit new terrorists by systematic brainwashing. [14] describes detection of influential users using linguistic approach. [15] uses graphs and networks (Social Networks Analysis -SNA) to detect leaders in the "dark networks".

Social Network Analysis is now widely used as a fully operational intelligence source, introducing the SOCMINT [27] neologism standing for Social Media Intelligence.

### 3.5 Knowledge Management for global security

Security management requires deep understanding of a given problem in its context and follows the rules of KM lifecycle: acquisition, modeling, navigation and use. The adequate AI techniques will support the whole process.

Global Security is composed of a set of domains security systems, each having its own objectives and controls. As threat awareness requires surveillance, each domain uses specific surveillance processes and vocabulary.

Knowledge Management helps organizing and optimizing the available sources of data/information and knowledge and creating progressively a common language for all involved actors to facilitate exchanges between them [33]. Three main approaches can be used: top-down, bottom-up and combination of both. In the case of global security a well-defined and dynamic strategy is mandatory to efficiently face the treats. Dynamic strategy allows flexibility for quick adaptation to evolving situation. Conceptual knowledge modelling as for ex ontologies allow reusing knowledge for related applications.

The existing literature provides description of all facets of security, useful for building such a system [25]. In this work, Rogers analyses the legacy of the Cold War's proliferation of weapons of mass destruction; the impact of human activity on the global ecosystem; the growth of hyper capitalism and resulting poverty and insecurity; the competition for energy resources and strategic minerals; biological warfare. This context leads to delinquency. Similar conclusion was stated at the end of colloquium organized by French Ministry of Defense on the influence of Climate change on the rise of crime [35].

Chen [27] describes Terrorism informatics as methodologies and integrate process of gathering and processing all related information.

We also found works applying KM principles to water and food security and to multiple risks management.

In our case Knowledge Management applied to global security system will connect components shown in Figure 1.

On our knowledge, there is no integrated approach of Knowledge Management to global security; this and the need for a systemic approach is leitmotiv for our research.

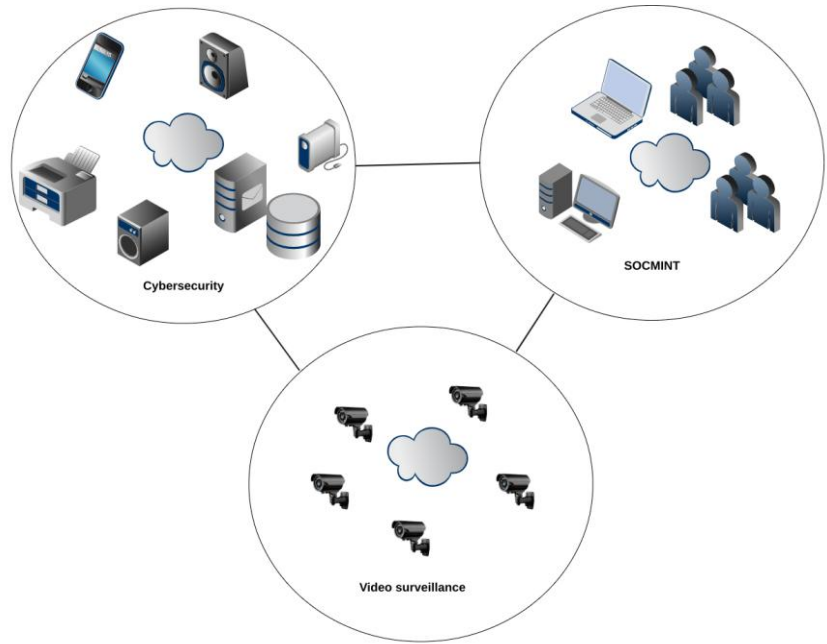

**Fig. 1.** Main interrelated components of global security system

Our proposal is then to consider the feedback from previous approach in the light of such global security system.

## 4      Feedback form the previous research

The research on video analysis, under the name of VORTEX project, has been initiated in 2010 in France after a large number of interviews with system architects, security managers and police officers. The aim was to spot the flaws of contemporary metropolitan video surveillance systems and propose an optimization of human activities in video surveillance operations. A synthetic article about public-private partnerships in new video surveillance services, for the French National Institute of Security and Justice (INHESJ), was written subsequently.
Findings described in this initial stage of the research where the following: As video surveillance moves through its digital transformation from analog cables, monitors and tapes to a complete computer based environment, there is  a quantum leap in both numbers of video sensors and geographic scale of systems deployed. Large French urban areas like Paris, Lyon, Marseille and Nice have set up or are in the process of setting up systems with more than one thousand cameras, without mentioning the thousands of cameras already scattered along the public transportation lines and inside the vehicles [1]. While the need for police activity monitoring surges, these numbers are also increased by the new body worn cameras on police officers. Information

Technology provides solutions to record and visualize all these cameras, but it does not meet the day to day exploitation needs made more complex by the multitude of video sensors. In a word, the capacity to have an eye everywhere does not spare the people watching. A global rethinking of the balance between people in front line, people in operations centers as well as citizens must be undertaken. We introduce the need for a rethinking and rationalization of the human role in image interpretation, based on the finding that we can deploy much more than we can actually monitor. It is made necessary to define how a human operator can collect and preserve intelligence [2] from video sources, with the aid of the machine, assuming the large number of video feeds creates a rich potential information source. This in turns requires new training procedures and new tools to be created to cope with system scale and carry out this strategic task.

Subsequent research emphasized the importance of a contextual use of video, in complement of other data sources.

Wise decision making require considering all available sources on a given topic as well as the knowledge about the current context. Recent attacks, notably on energy critical infrastructures [29] show that infrastructure security may be endangered by cyber-attacks. This is leading to the creation of a hybrid domain for security called Cyber Physical Security (CPS) [31] with social networks as platforms of choice for preparing operational details in such cases.

In this context, Vortex project objectives are to keep the human operator at the heart of the system and decision process. This requires the development of computer aided monitoring automation, providing advices and recommendations as to what should be watched first in the continuous flow of contextual real-time and recorded events.

## 5 Research focus – hybrid system for video analysis

Those breakthroughs in image and video analysis and labeling are the cornerstones of VORTEX concept. But as we witness the need for developing supervised machine learning processes that can lead to development of video intelligence expert modules, we also realize that in the ever changing very complex metropolitan environment, the patterns of normality and abnormality and their relationship to the images captured by the cameras are difficult to express. The role of the human operator in the heart of the semantic system is mandatory to reconcile volumes of data captured by computerized video sensors with contextual situation awareness. Reinforcement learning [31] is a machine learning technique sitting at the crossing of supervised and unsupervised learning. It implies defining a decision function that allows the machine to modify its state and also to define rewards and penalty functions that are used to provide the machine with an assessment of the result of its decisions. Using a very large number of try and fail the machine infers an optimized solution by maximizing its reward.

We therefore propose an approach based on two distinct annotation processes. One being conducted through the most modern labeling algorithms running on state of the art, dedicated hardware platforms, or inference platforms, the second being performed by the operators. We introduce a third knowledge based situational awareness module or recommendation module that uses insights produced from the analysis of combined human and machine generated annotations and communicates back its recommendations to the operator. This system is able of maintaining long term memory of what is a "normal" or "abnormal" situation and in addition has the essential capability to take into account human generated alerts and comments to adapt to new situations as they happen.

Replaced in the context of video surveillance, the human operator appears even more important to machine learning as he not only recognizes objects and people but assesses the level of risk of a particular situation and correlates scenes monitored by different cameras.

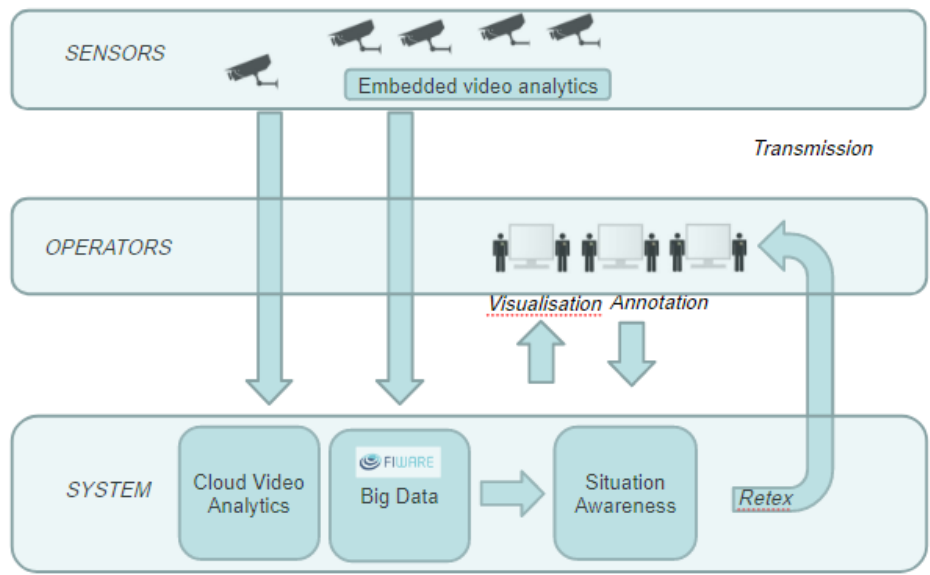

**Fig. 2.** Human Centric Design for man-machine interoperability

From the picture in fig. 2. it is made clear that the System is fed with annotations coming primarily from human operators' interactions witnessing events happening on their surveillance screens (traffic incidents, aggressions and thefts, vandalism, tags, terrorist attacks, smuggling…). This human input is key to providing a common sense context to information that is provided automatically by labeling modules either embedded directly in the cameras or located centrally in the cloud System.

The supervised learning is then operated by human rating of situations on a severity scale, enabling the System to learn and anticipate situations contexts leading to potential risky situations.

The RETEX1 describes that anticipation data which is deduced by the system while streams of new data flows continuously from both operators and cameras. These streams can also be completed by auxiliary data streams incoming from contextual communication systems and metadata concerning the sensors. The RETEX provides a predictive capacity based on the supervised learning achieved continuously by the interaction of operators and System. An important part of the System analytics is dedicated to transform the RETEX, primarily made of textual content, into actual operational data that can be actioned by operators. In the context of video surveillance, this is achieved by highlighting those cameras that are most important to watch.

### 5.1 Situational awareness increased with video surveillance

The RETEX illustrates how the predictive capabilities of the System can be turned into prescriptive surveillance actions. Still, state of the art video surveillance management system provide poor interfaces to enable operators to capture their annotations on the fly and store them in a workable format.

It is thus one of the key objectives of this research to study the conditions of an efficient real-time annotation to enable the operators to achieve the necessary supervised learning and initialize the RETEX loop. Information captured by operators are essential to a sound indexing of video and participate to the overall indexing required by both forensic investigations and day to day exploitation.

## 6 Originality of this work

VORTEX is proposing a genuine approach to the man machine cooperation by leveraging the recent breakthroughs in Machine Learning technologies that allows processing video from traditional stationary cameras as well as mobile or handheld devices [16]. Based on the essential finding that the video media needs to be translated to be workable, we propose to organize a data model for streaming information generated by video analytics labeling algorithms. We propose to define the annotation interfaces necessary to capture operators' annotation in real-time in such a way as it can be used as input for a supervised learning input. Eventually, we propose a predictive analytics System capable of issuing recommendations to the human operators and interacting with them in a feedback loop (RETEX) of reinforcement learning.

It is important to note that annotations as well as any other indexing data may be stored and preserved much beyond the limits of video retention periods, as advised for instance by the European Data Protection Supervisor (EDPS). This means that real-time automated labeling data are key to provide large datasets of surveillance contexts without the burden of keeping the video they originated from. VORTEX is an attempt to rationalize the capture of human feedback in surveillance and crisis context. Putting face to face these data with large volume of sensors data enable the System to corre-late and generalize on a rich dataset and leads to the emergence of predictive alerts and prescriptive surveillance actions that increase considerably the situational awareness in operations centers.

---

[1] RETEX is "Retour d'Experience" in French in the text (feedback from experience)

We are confronted with several difficulties, notably in scene description and modeling. But we are confident that we can circumvent those difficulties by using machine learning techniques rather than going through a scene data modeling exercise.

## 7 Conclusion

The described system is the first "Kblock[2]" of Decision support system for global security. The modular architecture will allow integration of other components such as conversations in social networks and data from IoT. Following the trend of predictive policing, we introduce a system that will help gathering intelligence from existing and future video surveillance systems and using it to anticipate terrorism and decrease safety risk in metropolitan areas.

Based on supervised Machine Learning and RETEX interaction loop, human operators will contribute to building System cognitive computing capacities and will be augmented in return by its prescriptive analytics.

Universality of this approach consists in the following: it does not depend on video surveillance technology infrastructure but complements it with new video analytics labeling systems, new annotation and communication tools and new predictive capabilities.

The security ontology definition is the basis of the underlying knowledge management required to provide a consistent framework that will serve as an interoperability guide to extend the approach to different countries and open intelligence cooperation between agencies, both nationally and internationally, representing a potential benefit for global organizations like EUROPOL. Moreover, the genericity of the model for risk management can be extended from video surveillance to cybersecurity to SOCMINT and to all surveillance activities.

Adopting an ontology and developing automated labeling capacities provides the ground for generating a continuous stream of data flowing from the many and highly diversified sources of information available, both video sensors and human inputs. Among human inputs we can cite metropolitan security control centers operators, but also social networks OSINT (Open Source Intelligence) or even SOCMINT (Social Media Intelligence) which play an ever increasing role in situational awareness. A mixt fusion approach based on cognitive computing, could then benefit large scale proven systems like IBM Watson [17] to extract early signals and anticipate risks from the very large textual information generated in such context.

---

[2] Knowledge block

## 8     Perspective and future work

VORTEX framework has been conceived for aiding urban video surveillance operations, but similar initiatives have been undertaken in the field of aerial image analytics [18] and the knowledge based information fusion proposed for the System has been under scrutiny in numerous other papers [19]. The range of sensors providing field data is not limited to stationary cameras. Ground vehicle cameras, aerial drone cameras, body-worn cameras, microphones and general presence detection sensors output information streams that can be injected in the RETEX interaction loop.

The real-time annotation tool may be utilized by operators supervising media different from urban surveillance cameras, i.e. thermal cameras, radars, LIDARs as well as front-line operators located directly in the zone of interest and providing direct field intelligence to the System.

Different application field also requiring human surveillance, like cyber security, may be using VORTEX framework by adapting the vocabulary of annotations. This is made possible by using domain dependent Ontologies, as mentioned in previous projects.

## 9     Acknowledgements

Inception of this research was presented in 2012 to the Aerospace competitiveness cluster PEGASE [22], now part of the larger "Pôle Risques" [23] in France where it got a distinction for its "usefulness in the aerial vehicles data processing allowing drones and stratospheric machines to achieve their mission".

## 10     Academic partnerships

VORTEX concept was elaborated in cooperation with two laboratories, the LIRMM from Montpellier University, expert in machine learning and the LUTIN from Paris VIII, specialized in man machine interfaces, detection and semantics of human perceived actions.